\documentclass[10pt, a4paper]{article}
\usepackage[final]{lrec-coling2024} 

\usepackage{natbib}
\usepackage{multibib}
\usepackage{tikz}
\makeatletter
\def\@mb@citenamelist{cite,citep,citet,citealp,citealt,citepalias,citetalias}
\makeatother
\newcites{languageresource}{~}

\usepackage{graphicx}
\usepackage{tabularx}
\usepackage{soul}
\usepackage{booktabs}
\usepackage{subfig}
\usepackage{titlesec}
\titleformat{\section}{\normalfont\large\bfseries\center}{\thesection.}{1em}{}
\titleformat{\subsection}{\normalfont\SmallTitleFont\bfseries\raggedright}{\thesubsection.}{1em}{}
\titleformat{\subsubsection}{\normalfont\normalsize\bfseries\raggedright}{\thesubsubsection.}{1em}{}
\renewcommand\thesection{\arabic{section}}
\renewcommand\thesubsection{\thesection.\arabic{subsection}}
\renewcommand\thesubsubsection{\thesubsection.\arabic{subsubsection}}

\usepackage{adjustbox}
\usepackage{xcolor}
\usepackage{hyperref}
 \definecolor{darkblue}{rgb}{0, 0, 0.5}
  \hypersetup{colorlinks=true, citecolor=darkblue, linkcolor=darkblue, urlcolor=darkblue}

\usepackage{xstring}

\usepackage{color}

\newcommand{\ie}{i.e.,~}
\newcommand{\eg}{e.g.,~}

\title{Enhancing Low-Resource LLMs Classification \\with PEFT and Synthetic Data}

\name{Parth Patwa$^1$, Simone Filice$^2$$^*$\thanks{$^*$Work done while at Amazon.}, Zhiyu Chen$^1$, \\ 
 {\bf \large Giuseppe Castellucci$^1$, Oleg Rokhlenko$^1$, Shervin Malmasi$^1$}
}

\address{$^1$Amazon, USA \\
        $^2$Technology Innovation Institute, Israel \\
         $^1$\{parthptw, zhiyuche, giusecas, olegro, malmasi\}@amazon.com, $^2$simone.filice@tii.ae\\}

\abstract{
Large Language Models (LLMs) operating in 0-shot or few-shot settings achieve competitive results in Text Classification tasks. In-Context Learning (ICL) typically achieves better accuracy than the 0-shot setting, but it pays in terms of efficiency, due to the longer input prompt. In this paper, we propose a strategy to make LLMs as efficient as 0-shot text classifiers, while getting comparable or better accuracy than ICL. Our solution targets the low resource setting, i.e., when only 4 examples per class are available. Using a single LLM and few-shot real data we perform a sequence of generation, filtering and Parameter-Efficient Fine-Tuning steps to create a robust and efficient classifier. Experimental results show that our approach leads to competitive results on multiple text classification datasets.
 \\ \newline \Keywords{LoRA, PEFT, LLMs, Few-Shot Learning, Text Classification
 } }

\begin{document}

\maketitleabstract

\section{Introduction}
\label{sec:intro}

Recent years have been characterized by a paradigm shift in text classification. Large Language Models (LLMs) offer valid alternatives to the traditional approach of fine-tuning pre-trained models (e.g., BERT \cite{devlin-etal-2019-bert}, RoBERTa \cite{liu2019roberta}) on annotated datasets. In-Context Learning (ICL) is a first option, where a LLM learns how to solve a task by simply observing a few examples provided in its prompt (i.e., without any fine-tuning stage) \cite{brown2020language}. Another alternative is the 0-shot setting, where the LLM directly solves a task by simply following the provided instructions (i.e., without any example). Instruction-fine tuned models like Flan-T5 \cite{chung2022scaling}, Instruct-GPT \cite{ouyang2022training}, or ChatGPT excel in this setting. The advantage of these two alternatives with respect to the traditional approach is that they can be used to bootstrap a model when data is scarce or totally absent.

The 0-shot setting is generally more appealing since it does not require any data, however, the few-shot ICL setting typically leads to better results by only leveraging a small number of samples (e.g., less than 10 examples). Obtaining a few annotated data might not be a significant drawback, as a practitioner with moderate domain knowledge can readily create a small number of examples manually. 
However, a major disadvantage is the higher computational cost, latency, and memory requirements associated with the longer prompt, which needs to contain illustrative examples.
A possible solution to leverage the few available examples without incurring the ICL inference costs would be to use them for fine-tuning the LLM, which is possible by using some Parameter-Efficient Fine Tuning (PEFT) techniques~\cite{liu_pt, lester-etal-2021-power, DBLP:journals/corr/abs-2106-09685, liu_ptv2}. Unfortunately, as we will demonstrate in the experimental section, PEFT is not effective with very few examples, due to under-fitting or over-fitting phenomena. 

In this paper, we propose a solution to the low-resource PEFT for text classification with LLMs by defining a framework that enables faster, cheaper and more accurate inference than ICL in such a scenario. We hypothesize that LLMs already have some knowledge of how to solve a classification task, but the sub-optimal usage of the available resources (i.e., the few-shot examples) results in low PEFT performance under the low-resource setting. On the contrary, LLMs typically excel in generation tasks, hence we frame an auxiliary data augmentation task that we use to unlock the LLM classification capabilities. Our method consists of three steps. First, we use the LLM to generate synthetic examples for each class of the text classification task we target. Then, we use the same LLM in the ICL setting to classify the examples and clean the data by removing label-inconsistent generated examples. Finally, we fine-tune the LLM with PEFT using the generated and cleaned data. Our experiments show that the resulting classifier reaches accuracy levels comparable to or better than the ICL setting in three different text classification tasks while being a lot more efficient ($\sim$2x to 5x speed boost). 
In these generate-filter-train stages we always use the same LLM to demonstrate that what leads to a good accuracy is just a better usage of the few available examples and not the employment of any other resource (e.g., another LLM) which might bring additional knowledge to solve the task.

The rest of the paper is organized as it follows: in Section \ref{sec:related} we discuss the related works. In Section \ref{sec:method} we present our method, while in Sections \ref{sec:exp} and \ref{sec:results} we discuss the experimental setting and the results, respectively. Finally, Section \ref{sec:conclusions} derives the conclusions.
\begin{figure}[t!]
    \centering
    \includegraphics[width=\linewidth]{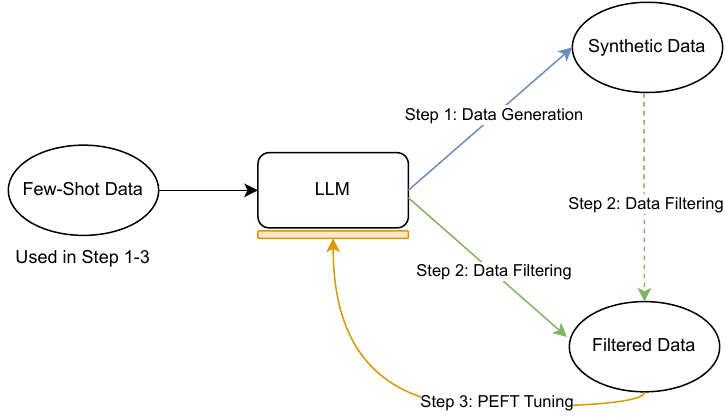}
    \caption{The overview of our method. First, very few real data points are used to generate synthetic data using ICL. Then, the synthetic data is filtered using ICL by LLM again. Finally, the filtered data and the real data are combined to train the LLM using LoRA. 
    }
    \label{fig:flow}
\end{figure}
\section{ Related Work}
\label{sec:related}

LLMs demonstrate impressive capabilities to solve Natural Language Understanding tasks. For instance, classification tasks can be approached in a generative way, \ie by asking the LLM to generate the class name associated with an input example. 0-shot and ICL are two variants of this paradigm. 
Recent models (\eg GPT-3, \cite{brown2020language}, Flan-T5 \cite{chung2022scaling}, ChatGPT \cite{ouyang2022training}, Falcon \cite{penedo2023refinedweb} or Vicuna \cite{Vicuna}) reach impressive results in both settings, and researchers started considering using LLMs as annotators~\cite{rosenbaum-etal-2022-linguist, zhu2023can, he2023annollm}. For instance, \citet{rosenbaum-etal-2022-linguist} propose a method that uses LLMs to generate and annotate data for Intent Classification and Slot Filling. \citet{he2023annollm} propose a two-step approach where they first use ChatGPT to generate a few-shot Chain-of-Thought prompt, which they then use to annotate unlabeled data. Results are competitive with human annotators, but their classification procedure is relatively slow since the LLM is invoked twice with prompts that need to contain both examples and explanations. Conversely, we propose a solution whose computational complexity at inference time corresponds to the 0-shot setting case.
Even if these results are impressive, they still might not reach the state-of-the-art performance achievable when LLMs are fully fine-tuned on large data. 
Fine-tuning LLMs is extremely expensive, but a viable solution is offered by Parameter Efficient Fine-Tuning (PEFT) techniques \cite{liu_pt, lester-etal-2021-power, DBLP:journals/corr/abs-2106-09685, liu_ptv2}, where a pre-trained model is fine-tuned by only updating a small number (e.g., 0.01\%) of added or selected parameters. These methods report results that match the performance of full fine-tuning when large training datasets are available. On the contrary, there has been relatively little focus on (parameter-efficient) fine-tuning in low-resource settings. Our paper targets this scenario, as we assume we can access only a few annotated examples (\eg four per class) and no unlabeled data. A work operating in a similar setting is \citet{NEURIPS2022_0cde695b}, where the authors propose a novel PEFT technique that is demonstrated to work well in low resource settings when the PEFT weights are pre-trained and multiple tasks are trained in parallel. We differ from their work as we do not pre-train the PEFT weights and we target a single task at a time, without assuming (possibly related) data from other tasks is available. Relaxing this assumption is especially useful when dealing with very peculiar tasks not sharing similarities with other available datasets. Hence, to the best of our knowledge, we are the first to improve few-shot PEFT without additional resources (external datasets or models).

\section{Methodology }
\label{sec:method}

\begin{table*}[ht!]
\centering
\resizebox{2\columnwidth}{!}{%
\begin{tabular}{|l|l|l|l|ll|ll|ll|}
\hline
\textbf{Model}  & \textbf{Method} & \textbf{\#real} & \textbf{\#syn} & \multicolumn{2}{c|}{\textbf{SST2}}             & \multicolumn{2}{c|}{\textbf{TREC}}             & \multicolumn{2}{c|}{\textbf{AG News}}           \\ \hline
       &        &         &        & \multicolumn{1}{c|}{\textbf{acc}}  & \textbf{inf. time} & \multicolumn{1}{c|}{\textbf{acc}}  & \textbf{inf. time} & \multicolumn{1}{c|}{\textbf{acc}}  & \textbf{inf. time} \\ \hline
Vicuna7b  & 0-shot    & 0       & 0      & \multicolumn{1}{l|}{0.55} & 0.27       & \multicolumn{1}{l|}{0.16} &     0.28   & \multicolumn{1}{l|}{0.36} &    0.5    \\ 
Vicuna7b  & ICL    & 4       & 0      & \multicolumn{1}{l|}{\textbf{0.95}} & 0.6       & \multicolumn{1}{l|}{0.60} & 0.9       & \multicolumn{1}{l|}{0.75} & 2.5       \\ 
Vicuna7b  & LoRA   & 4       & 0      & \multicolumn{1}{l|}{0.51} & 0.27      & \multicolumn{1}{l|}{0.49} & 0.28      & \multicolumn{1}{l|}{0.35} & 0.5       \\ 
Vicuna7b  & LoRA   & 25      & 0      & \multicolumn{1}{l|}{0.89} & 0.27      & \multicolumn{1}{l|}{\textbf{0.84}} & 0.28      & \multicolumn{1}{l|}{\textbf{0.86}} & 0.5       \\
Vicuna7b  & ours  & 4       & 21     & \multicolumn{1}{l|}{0.90} & 0.27      & \multicolumn{1}{l|}{0.79} & 0.28      & \multicolumn{1}{l|}{0.82} & 0.5       \\ \midrule
Vicuna13b & 0-shot    & 0       & 0      & \multicolumn{1}{l|}{0.85} & 0.37      & \multicolumn{1}{l|}{0.36} & 0.38       & \multicolumn{1}{l|}{0.31} & 1.83     \\ 
Vicuna13b & ICL    & 4       & 0      & \multicolumn{1}{l|}{0.93} & 1.2       & \multicolumn{1}{l|}{0.75} & 1.7       & \multicolumn{1}{l|}{0.80} & 4.36      \\ 
Vicuna13b & LoRA   & 4       & 0      & \multicolumn{1}{l|}{0.84} & 0.37      & \multicolumn{1}{l|}{0.62} & 0.38      & \multicolumn{1}{l|}{0.64} & 1.83      \\ 
Vicuna13b & LoRA   & 25      & 0      & \multicolumn{1}{l|}{0.93} & 0.37      & \multicolumn{1}{l|}{\textbf{0.93}} & 0.38      & \multicolumn{1}{l|}{0.84} & 1.83      \\ 
Vicuna13b &  ours  & 4       & 21     & \multicolumn{1}{l|}{\textbf{0.93}} & 0.37      & \multicolumn{1}{l|}{0.81} & 0.38      & \multicolumn{1}{l|}{\textbf{0.86}} & 1.83      \\ \bottomrule
Vicuna7b  & LoRA  & Full       & 0    & \multicolumn{1}{l|}{0.97} & 0.27      & \multicolumn{1}{l|}{0.98}    & 0.28         & \multicolumn{1}{l|}{0.95}    & 0.5         \\ \bottomrule
\end{tabular}%
}
    \caption{Accuracy results and inference times (in seconds) on three classification tasks. In bold the best performing method for the model in a data-scarce setting. "Full" data refers to the use of the entire available training data.  \#real and \#syn refer to the number of real and synthetic examples per class used for training.}
    \label{tab:all-results}
\end{table*}

We explore a low-resource setting where we have few training examples per class and no unlabeled data.  ICL methods could achieve reasonable performance with few-shot samples but inference cost is high due to long prompts. PEFT methods like LoRA are known to be more efficient than ICL at inference. However, we find that LoRA performs worse than ICL in data-scarce settings (see Tab. \ref{tab:all-results}). In this paper, we aim to explore the potential of combining the strengths of PEFT and ICL methods for achieving efficient and effective text classification.
Hence we propose to augment the training data with synthetic data to better align the generation and classification capability of LLMs and to ensure that PEFT is performed on a decent amount of data. Our method has 3 steps: generate data, filter data, and train. An overview of our method is shown in Figure \ref{fig:flow}. 

\begin{figure}[t]
    \centering
    \begin{tikzpicture}
    \node[draw, rounded corners] {
    \resizebox{0.75\linewidth}{!}{    
    \adjustbox{minipage=[r][13em][b]{0.4\textwidth},scale={0.7}}{
    \textbf{Few examples of movie reviews having positive sentiment are given. Generate more positive reviews}\\
    \textbf{\textcolor{blue}{Text:}} [Positive review 1] \\
    \textbf{\textcolor{blue}{Label:}} Positive\\
     \textbf{\textcolor{blue}{...}} \\
    \textbf{\textcolor{blue}{Text:}} [Positive review 4]\\
    \textbf{\textcolor{blue}{Label:}} Positive\\
    \textbf{\textcolor{blue}{Text:}} \textcolor{orange}{[the model generates this]}\\
    }}
    };
    \end{tikzpicture}
    \caption{An example of a prompt used for generating positive reviews for SST2 data. Four examples of the positive class are provided in the prompt. }
    \label{fig:prompt_generate}
\end{figure}

\begin{figure}[t!]
    \centering
    \begin{tikzpicture}
    \node[draw, rounded corners] {
    \resizebox{0.75\linewidth}{!}{    
    \adjustbox{minipage=[r][12em][b]{0.4\textwidth},scale={0.7}}{
    \textbf{Classify the sentiment of the given movie review into Positive or Negative}\\
    \textbf{\textcolor{blue}{Text:}} [review 1] \\
    \textbf{\textcolor{blue}{Label:}} [Label 1]\\
     \textbf{\textcolor{blue}{...}} \\
    \textbf{\textcolor{blue}{Text:}} [review 8]\\
    \textbf{\textcolor{blue}{Label:}} [Label 8]\\
    \textbf{\textcolor{blue}{Text:}} \textcolor{violet}{[generated review]}\\
    \textbf{\textcolor{blue}{Label:}} \textcolor{orange}{[Predicted Label]}
    }}
    };
    \end{tikzpicture}
    \caption{An example of a prompt used for classifying the sentiment of a movie review. Four examples per class are given in the prompt in a random order.}
    \label{fig:prompt_classify}
\end{figure}

\noindent
\textbf{Generation step:} \citet{chavan2023oneforall} show that in a few-shot setting, the performance of PEFT significantly improves as the number of training samples per class increases. We also observe similar results in our initial experiments (presented in section \ref{sec:results}.  Further, since ICL performs well, we hypothesize that the model has the inherent knowledge to solve the classification task and that the low PEFT results are due to sub-optimal usage of the available resources (the few shot examples). To fill this gap, we first use the LLM $\mathcal{L}$ in the ICL setting to generate synthetic data which we can use to augment the few shot examples at our disposal. We generate examples for each class in the targeted classification task. An example of the prompt we used in this step is shown in Figure \ref{fig:prompt_generate}.

\noindent
\textbf{Filtering step:} We first apply a basic filtering step to discard duplicates and malformed generations (i.e., too short or too long texts). On manual inspection of the generated data, we found some label-inconsistent generations (i.e., data that are not valid examples of the class they should represent). We hypothesize this is due to hallucination. To identify and remove these cases, we classify the generated data using ICL with $\mathcal{L}$. The prompt used for this stage is similar to the one shown in Figure \ref{fig:prompt_classify}. If the predicted label does not match the intended label from the generation step, we discard the generated example. We repeat these generation-filtering steps until we produce \textit{N} new data samples for each class in the targeted classification task.

\noindent
\textbf{Training step:} Finally, we use the filtered data along with the few (4 per class) real examples for the PEFT of the LLM $\mathcal{L}$ with LoRA. Note that $\mathcal{L}$ is used for all 3 steps, as we want to validate our hypothesis that $\mathcal{L}$ does not need additional knowledge to work in the PEFT setting, but only a more stable training process which can be guaranteed by the self-generated synthetic examples. 

\noindent \textbf{Inference} Conversely to the ICL setting, the LLM does not use any example at inference time. Note that the three steps (generate, filter, train) are used only at training time, i.e., there is no impact on the inference latency.

\begin{figure*}%
    \centering
    \subfloat[\centering Real data]{{\includegraphics[width=0.5\linewidth]{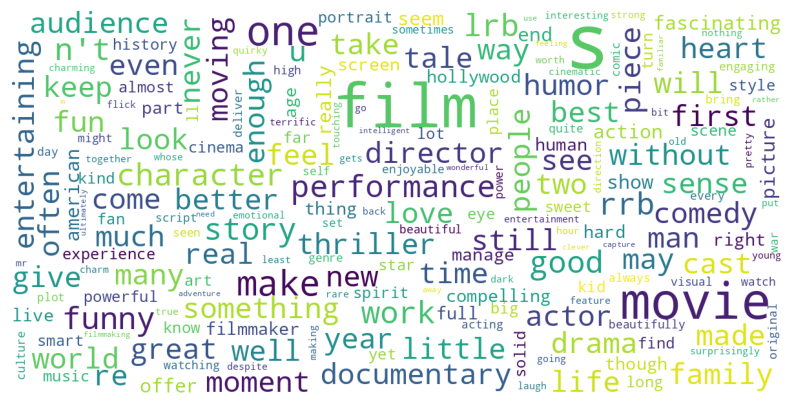} }}%
    \subfloat[\centering Synthetic data]{{\includegraphics[width=0.5\linewidth]{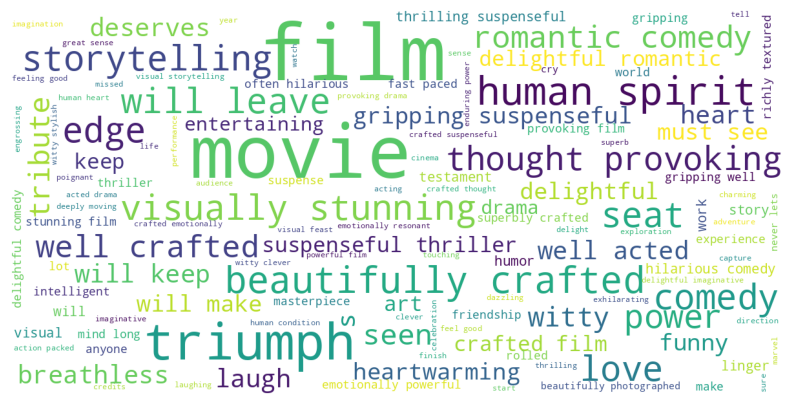} }}%
    \caption{Word clouds of the real and synthetic data belonging to the positive class in SST2. }%
    \label{fig:pos-wc}%
\end{figure*}

\section{Experiments}
\label{sec:exp}

We use the Vicuna LLM \cite{Vicuna}, which is based on LLaMA \cite{llama}. We selected Vicuna as it is the best-performing model among those whose weights were publicly available at the time of our experiments, according to the Chatbot Arena Leaderboard\footnote{\url{https://lmsys.org/blog/2023-05-25-leaderboard/}}. We experiment with 2 model sizes - Vicuna-7b and Vicuna-13b which have 7 billion and 13 billion parameters, respectively. We show results on the official test sets of 3 datasets -  SST2 (sentiment analysis, 2 classes) \cite{sst2}, AG News (news topic classification, 4 classes) \cite{agnews}, and TREC (question classification, 6 classes) \cite{trec}.  
To generate synthetic data, we use random sample decoding with temperature = 1.0, top\_k = 50 and num\_beams = 1 or 2.

For a fair comparison across models and methods, throughout our experiments, we do not tune the LoRA hyper-parameters (we set rank=8, alpha=32, and dropout=0.1). Similarly, we make minimal changes to the prompt for all ICL experiments and do not perform any prompt engineering.

The code is implemented using hugging face and PEFT library \cite{peft} with torch backend. All models are trained on 4 v100 GPUs of 16GB each. All LoRA models are trained for 100 epochs. The language modeling loss (next token prediction) is optimized using the Adam optimizer. Batch size is set to 2 for Vicuna-13b and 8 for Vicuna-7b. We run each experiment multiple times with different seeds and different few-shot examples. We observe negligible deviation across runs and report the average accuracy. We also report the results of several baselines, including 0-shot, ICL, and vanilla LoRA trained with different numbers of real examples. We also report LoRA trained with the full training set. This could be considered a potential upper-bound reachable in high resource settings and gives results comparable to the respective SoTA methods on the 3 datasets \cite{raffel2023exploring,cer2018universal,xlnet}.

\section{Results and Analysis}
\label{sec:results}

Table \ref{tab:all-results} reports the accuracy results and inference times on three tasks. As expected, the few-shot ICL is more accurate than the 0-shot method but significantly slower.  Across model sizes and datasets, vanilla LoRA performs a lot worse than ICL in a few shot setting (4 real data points per class). Furthermore, it is even worse than or comparable to the 0-shot method in some cases. Conversely, our generate-filter-train approach is comparable to ICL on the SST2 data and considerably outperforms the ICL baseline on the other datasets. For example, when using Vicuna-7b model on the TREC data, our method gives 0.84 accuracy whereas ICL gives only 0.6 accuracy. Our method is also comparable to or slightly worse than training LoRA with 25 real samples per class. Note that the inference time of LoRA is lower and independent of training data size, whereas the inference time of few-shot ICL is higher and increases with an increase in training data \cite{ia3}. As expected, the overall performance is generally better with Vicuna-13b than with Vicuna-7b.

\noindent
\textbf{Ablation studies:} We train LoRA on the unfiltered synthetic data (by skipping the filtering step) and find that it achieved an accuracy of 0.68 on the TREC dataset. Using the filtered data the accuracy is 0.79, which shows the usefulness of the filtering step. We manually checked 125 unfiltered synthetic TREC samples and we found that 33 were incorrect (cases of hallucination). For example \textit{"Who is called the Father of Geometry”} was incorrectly generated for the \texttt{location} class. However, only 5 samples were incorrect out of 125 examples after filtering.  For the SST2 dataset, no sample was identified as incorrectly labeled, hence the filtering step did not affect the performance.

\begin{figure}[t]
    \centering
    \includegraphics[width=\linewidth]{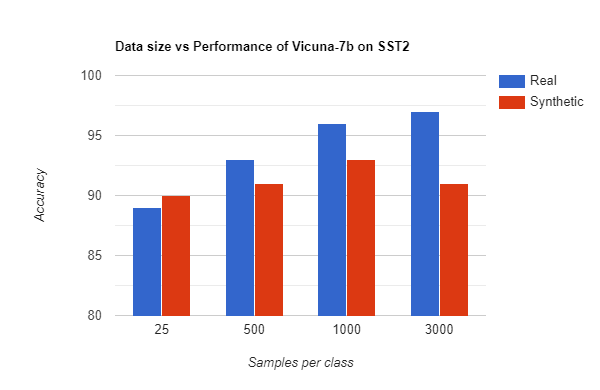}
    \caption{Data size vs performance of Vicuna-7b on SST2.}
    \label{fig:datasize-sst2}
\end{figure}

\begin{figure}[t]
    \centering
    \includegraphics[width=\linewidth]{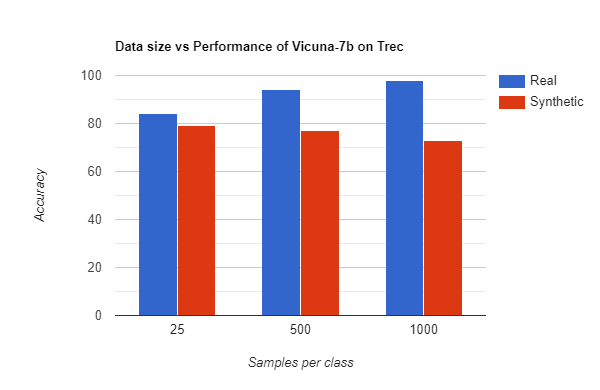}
    \caption{Data size vs performance of Vicuna-7b on TREC. Please note that in the real dataset, some classes have a limited number of instances (possibly lower than the values reported on x-axis). In such case, all instances of these classes are used.}

    \label{fig:datasize-trec}
\end{figure}

\noindent
\textbf{Effect of Data Size:} We experiment with various data sizes. The results of Vicuna-7b on SST2 and TREC are shown in Figure \ref{fig:datasize-sst2} and \ref{fig:datasize-trec}, respectively. Increasing the size of real data is always beneficial; on the opposite, adding more synthetic data does always not provide a clear benefit. The main reason is the lack of diversity in the synthetic data.

\begin{figure}[t]
    \centering
    \includegraphics[width=0.9\linewidth]{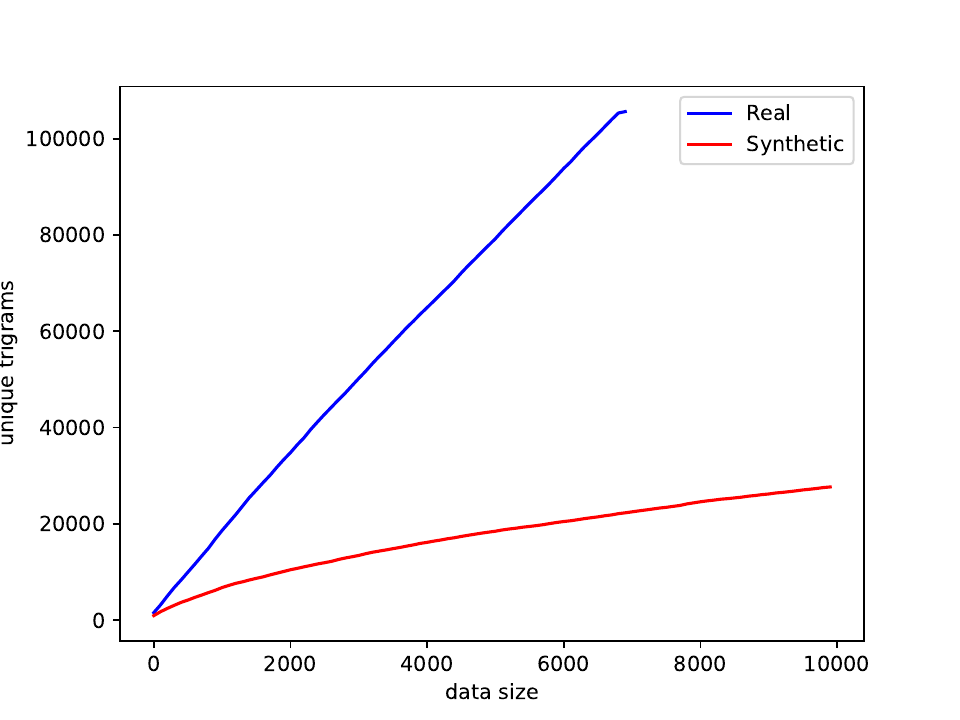}
    \caption{Data size vs unique trigrams in SST2. }
    \label{fig:sizevstri}
\end{figure}

\noindent
\textbf{Data diversity:} Figure \ref{fig:sizevstri} shows the number of unique tri-grams vs. data size for real and synthetic SST2 data. We can see that for smaller data sizes, the diversity of the real data is comparable to that of synthetic data. However, as the data size increases, the real data diversity increases faster. This shows the difficulty of generating a large amount of diverse synthetic data with only 4 seed examples.

\noindent
\textbf{Qualitative Data Analysis:} Figure \ref{fig:pos-wc} shows the word clouds of the real and synthetic examples belonging to the \textit{positive} class of SST2 data. SST2 is a dataset of sentiment analysis of movie reviews. Hence, we can see that words like \textit{film}, \textit{movie}  appear in both word clouds. Words that show positive sentiment like \textit{entertaining}, \textit{beautiful}, \textit{funny} also appear in both the word clouds. Further, we see other positive words like \textit{stunning}, \textit{delightful} only in the synthetic data word cloud whereas subtle positive words like \textit{compelling}, \textit{solid} are seen only in the real data word cloud. From this, we can conclude that the synthetic data has a slightly different distribution and can capture the meaning of the positive class. 

\section{Conclusion and Future work}
\label{sec:conclusions}
In this paper, we introduced a framework to make LLMs more efficient and effective text classifiers in very low-resource settings. The procedure we proposed consists of three steps. In the first step, the LLM is used to augment a very small training set with synthetic data; then, we adopt the LLM to classify the generated data and remove label-inconsistent examples; finally, we use the resulting data to fine-tune the LLM using LoRA. By running experiments on three different classification datasets we demonstrated how training LoRA using the self-generated synthetic data allowed our model to be comparable to or surpass several baselines operating in low resource settings, including 0-shot, ICL, and vanilla LoRA. In future work, we plan to improve the quality of the generated examples by promoting data diversity. Some strategies to improve data diversity include increasing attribute diversity \cite{diversity1}, logit suppression \cite{diversity2} etc. 

\section{Limitations}
Our method might not work on tasks that are particularly challenging and hard to catch with only a few examples. In this case, ICL is expected to fail, and similarly, our first two steps are expected to produce low-quality examples making the entire procedure ineffective. Another limitation is that our approach is fully based on LLMs and cannot be applied to low-resource languages where there is no existing LLM working well.

\section{Ethics}
Generating data using LLMs for text classification exposes the resulting classifier to the biases acquired during the LLM pre-training. In our framework, this phenomenon is potentially even amplified, as using the same LLM to generate and filter the data might reinforce such biases. Unfortunately, there is no one-size-fits-all solution for this problem. The biases are dependent on the application domain and on the data distribution to be generated. However, we encourage the readers to be very cautious about using this framework and to take the appropriate actions - for example, compiling a list of the potential biases specific for the target application domain and checking for those in the generated data -  to mitigate the potential biases that may get reinforced when using a methodology similar to the one here presented.

\section{References}\label{reference}

\bibliographystyle{lrec-coling2024-natbib}
\bibliography{lrec-coling2024-example}

\end{document}